# Casting Robotic End–effectors To Reach Faraway Moving Objects

Adriano Fagiolini, Hitoshi Arisumi, and Antonio Bicchi

*Abstract*—In this article we address the problem of catching objects that move at a relatively large distance from the robot, of the order of tens of times the size of the robot itself. To this purpose, we adopt casting manipulation and visual–based feedback control. Casting manipulation is a technique to deploy a robotic end-effector far from the robot's base, by throwing the end–effector and controlling its ballistic flight using forces transmitted through a light tether connected to the end–effector itself. The tether cable can then be used to retrieve the end-effector to exert forces on the robot's environment.

In previous work, planar casting manipulation was demonstrated to aptly catch static objects placed at a distant, known position, thus proving it suitable for applications such as sample acquisition and return, rescue, etc. In this paper we propose an extension of the idea to controlling the position of the end-effector to reach moving targets in 3D. The goal is achieved by an innovative design of the casting mechanism, and by closing a real–time control loop on casting manipulation using visual feedback of moving targets. To achieve this result, simplified yet accurate models of the system suitable for real–time computation are developed, along with a suitable visual feedback scheme for the flight phase. Effectiveness of the visual feedback controller is demonstrated through experiments with a 2D casting robot.

## I. INTRODUCTION

In several robotics applications, such as planetary exploration missions involving material sample acquisition and return, the possibility of reaching large workspace would afford great potential advantages. To operate on objects at distances several times larger than the physical dimensions of the robot, mobile platforms [1] equipped with articulated arms are practically the only available solution at the state of the art. However, wheeled or legged robotic locomotion depends heavily on the characteristics of the terrain, and it is usually forced to trade velocity of execution for robustness to terrain asperities. As a matter of fact, e.g., Martian explorers Spirit and Opportunity have been traveling at max. 180 m/h speed, on an average mission length of 100 meters from the base station, thus limiting the number of samples returned per day. On the other hand, the alternative of building arms with either very long links [2] or many links [3], [4] seems to be applicable only in some very specific cases — for instance in the absence of gravity — and yet imposes the use of very wide mechanical

A. Fagiolini is with the Interdepartmental Research Center "E. Piaggio" of the Università di Pisa, Italy, a.fagiolini@ing.unipi.it.
H. Arisumi is with the Intelligent Systems Research Institute, AIST, Tsukuba, Japan, h-arisumi@aist.go.jp.
A. Bicchi is with the Interdepartmental Research Center "E. Piaggio" of the Università di Pisa, Italy, bicchi@ing.unipi.it.

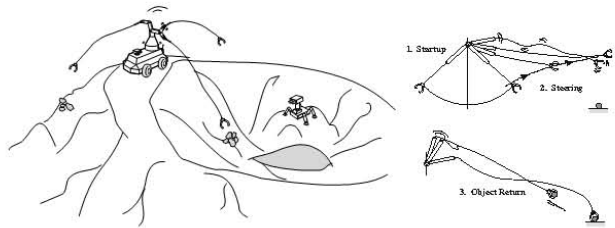

Figure 1. A robotic end–effector connected to a manipulator through a light tether can be used to catch far away from the manipulator base.

structures despite the extension of their reachable spaces. By contrast a casting manipulator can easily reach several meters from its base on Earth, and an even longer distance in low gravity.

In this paper we present work aimed at developing a compact robotic device able to reach objects at far distance. The work is based on the *casting manipulation* technique, proposed first in [5] and allowing the deployment of an end–effector at large distances from the robot's base by throwing (casting) it and controlling its ballistic flight using forces transmitted through a light tether connected to the end–effector itself. The tether cable of the robotic device can also be used to retrieve the end–effector and to exert forces on the robot's environment. The operating phases of casting manipulation comprise a *startup phase*, a *steering phase*, and an *retrieval phase*. During the startup phase, the robot is controlled so as to impart the end–effector sufficient mechanical energy to reach the target–object. When the first phase concludes, the end–effector is thrown, and its trajectory is steered by means of forces transmitted through the tether cable in order to approach the moving object with suitable orientation and velocity (steering phase). Once the object has been caught, the tether cable is reeled up and the object is retrieved (object–return phase). A possible application scenario for casting manipulation in a sample–return mission is shown in Fig. 1 along with a depiction of the different operating phases of the technique.

Ability of planar casting manipulators to fetch distant objects placed at *a priori* known positions has been demonstrated in [6]–[9] with early hardware prototypes. In [9], so–called *multiple braking control* was proposed as a technique to steer the robot's gripper toward a target–object, by applying impulsive forces to the gripper itself. By doing so, it was shown that a fixed object placed on the throwing plane can indeed be reached with suitable gripper's orientation and velocities. We propose an extension of the idea to controlling the position of the end–effector in order to reach moving objects in 3D. The goal is achieved by an innovative design of the casting mechanism and by closure of a real–time control loop on casting manipulation that uses visual feedback of moving targets.

Furthermore, reaching or better yet catching moving objects poses strict real–time specifications on the computation time of the inherent control strategy. This makes the *real–time casting manipulation problem*, including 3D control of the end–effector position, orientation, and velocity, a formidable challenge. For this reason, we use simplified yet accurate models for the steering phase that are based on a form of visual feedback, but still remain amenable for real–time computation. More precisely, we propose two visual–based controllers for the steering phase of a 2D casting manipulator: the former is a variation of the impulse–based approach proposed in [9] for the mobile target case, and the latter is a more general approach that consists of steering the end–effector by exertion of piecewise–constant forces through the tether cable on the end–effector itself.

The extension to the 3D case concerns the design of a new casting mechanism. Other works on cable–driven robots study only their steady state configuration, whereas our work's novelty is represented also by the investigation of the robots' dynamics [10]. Indeed, to achieve good performance, we focus on an optimal control problem requiring that the end–effector is steered to a target position in minimum–time. Optimization also concerns the value of the throwing angle w.r.t. the direction of the target–object.

In conclusion, the paper contains two major contributions: the visual–based feedback controllers that allow us to reach moving objects, and the extension of the casting technique to 3D control of the robotic end–effector's flight. The former result is validated through experiments, whereas the latter one is theoretically studied on a prototypical 1–cable–driven robot and then evaluated by simulation. The paper, based in part on the authors' previous work in [11] and [12], is organized as follows. Sec. II deals with modeling and control of a planar casting manipulator and presents two control methods suitable for real–time planning of the end–effector's trajectory. Then, Sec. III deals with the control of the innovative mechanism for 3D casting. Sec. IV describes the experimental setup consisting of a vision and control system, and it compares the two methods through the results obtained with a 2D planar casting manipulator. Finally, work achievements are summarized in Sec. V.

## II. PLANAR CASTING MANIPULATION

### A. Modeling and control of the robot

Consider the robot depicted in Fig. 2 which can be used for catching objects on a plane. The robot consists of a rigid link $L_1$ with revolute joint $q_1$ actuated by torque $\tau_1$, a tether cable departing from $L_1$, and a gripper of mass $m_3$ attached to the cable itself. Angle $q_2$ at which the tether departs from the rigid link is measured but not actuated. The cable is winded around a reel controlled by force $f_3$, and it can be considered as an actuated translational joint $q_3$. Let $a_1$ be the center of mass of the first joint.

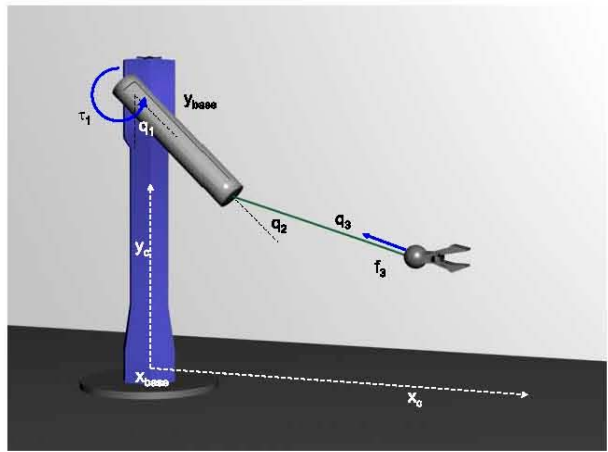

Figure 2. Depiction of a planar casting manipulator consisting of a rigid link, a light tether, and a gripper.

Flexibility of the tether cable must be taken into account to appropriately model the system dynamics. However, a simplified yet accurate model of the startup and steering phases can be found, if a suitable control strategy is adopted guaranteeing that the tether never becomes loose, and elastic modes are never excited (cf. e.g. [13]–[15]). If so, the behavior of the tether cable can be approximated as that of a rigid link, and the robot's dynamics can be conveniently written in the classical form:

$$\mathrm{B}(q)\ddot{q} + \mathrm{C}(q,\dot{q})\dot{q} + \mathrm{G}(q) = \tau, \qquad (1)$$

where $\mathrm{B}, \mathrm{C} \in \mathbb{R}^{3\times 3}$, $\mathrm{G} \in \mathbb{R}^3$, $q = (q_1, q_2, q_3)^T$, and $\tau = (\tau_1, 0, -f_3)^T$. Elements of matrices B, C, and vector G are reported in Table I, where the following standard abbreviations are used: $\tilde{C}_i = \cos(q_i)$, $\tilde{S}_i = \sin(q_i)$, $\tilde{C}_{ij} = \cos(q_i + q_j)$, and $\tilde{S}_{ij} = \sin(q_i + q_j)$.

It is worth remarking that the tether cable can exert only pulling force on the end–effector, and hence the control input $f_3$ must not be negative, i.e. $f_3(t) \geq 0$. Another important concern is represented by the fact that the manipulator is underactuated: in fact, there are two control inputs, $\tau_1$ and $f_3$, and three joint variables, $q_1$, $q_2$, and $q_3$. Then, as in [16], we will adopt a control strategy by which a subset of joints are steered so as that the remaining ones follow a desired trajectory. For this aim, let us define the system output as $\zeta = (q_2, q_3)^T$. In the remainder of this section, we will show that the system dynamics of Eq. 1 can be exactly linearized w.r.t. $\zeta$ by a suitable feedback law applied through controls $\tau_1$ and $f_3$. In our case, since the relative degree of the system is 4, such a linearization can be only partial. However, we show that the remaining dynamics, namely so–called zero–dynamics, is asymptotically stable.

To show this result, let us first left–multiply the dynamics' relation in Eq. 1 by the inverse of the robot inertia matrix, $\mathrm{B}(q)^{-1}$, thus obtaining:

$$\ddot{q} = \mathrm{B}(q)^{-1}\left(\tau - \mathrm{C}(q,\dot{q})\dot{q} - \mathrm{G}(q)\right),$$

Table I
ELEMENTS OF MATRICES B, C, AND G OF THE SIMPLE PLANAR CASTING MANIPULATOR.

$$
\begin{aligned}
b_{11} &= I_1 + I_3 + m_1 l_1^2 + \\
&\quad + m_3 \left(a_1^2 + q_3^2 + 2 a_1 q_3 \tilde{C}_2\right) \\
b_{12} &= I_3 + m_3 \left(q_3^2 + a_1 q_3 \tilde{C}_2\right) \\
b_{13} &= m_3 a_1 \tilde{S}_2 \\
b_{21} &= b_{12} \\
b_{22} &= I_3 + m_3 q_3^2 \\
b_{23} &= 0 \\
b_{31} &= b_{13} \\
b_{32} &= b_{23} \\
b_{33} &= m_3 \\
c_{11} &= -m_3 a_1 q_3 \tilde{S}_2 \dot{q}_2 + m_3 \left(a_1 \tilde{C}_2 + q_3\right) \dot{q}_3 \\
c_{12} &= -m_3 a_1 q_3 \tilde{S}_2 \left(\dot{q}_1 + \dot{q}_2\right) + m_3 \left(a_1 \tilde{C}_2 + q_3\right) \dot{q}_3 \\
c_{13} &= m_3 \left(a_1 \tilde{C}_2 + q_3\right) \left(\dot{q}_1 + \dot{q}_2\right) \\
c_{21} &= m_3 a_1 q_3 \tilde{S}_2 \dot{q}_1 + m_3 q_3 \dot{q}_3 \\
c_{22} &= m_3 q_3 \dot{q}_3 \\
c_{23} &= m_3 q_3 \left(\dot{q}_1 + \dot{q}_2\right) \\
c_{31} &= -m_3 \left(a_1 \tilde{C}_2 + q_3\right) \dot{q}_1 - m_3 q_3 \dot{q}_2 \\
c_{32} &= -m_3 q_3 \left(\dot{q}_2 + \dot{q}_3\right) \\
c_{33} &= 0 \\
g_1 &= m_1 g l_1 \tilde{S}_1 + m_3 g \left(a_1 \tilde{S}_1 + q_2 \tilde{S}_{12}\right) \\
g_2 &= m_3 g q_3 \tilde{S}_{12} \\
g_3 &= -m_3 g \tilde{C}_{12}
\end{aligned}
$$

which can be explicitly written as

$$
\begin{pmatrix} \ddot{q}_1 \\ \ddot{q}_2 \\ \ddot{q}_3 \end{pmatrix} = \begin{pmatrix} b_{11} & b_{12} & b_{13} \\ b_{21} & b_{22} & b_{23} \\ b_{31} & b_{32} & b_{33} \end{pmatrix}^{-1} \begin{pmatrix} \gamma_1 + \tau_1 \\ \gamma_2 \\ \gamma_3 - f_3 \end{pmatrix}, \quad (2)
$$

where

$$
\gamma_1 = -c_{11} \dot{q}_1 - c_{12} \dot{q}_2 - c_{13} \dot{q}_3 - g_1,
$$
$$
\gamma_2 = -c_{21} \dot{q}_1 - c_{22} \dot{q}_2 - c_{23} \dot{q}_3 - g_2,
$$
$$
\gamma_3 = -c_{31} \dot{q}_1 - c_{32} \dot{q}_2 - c_{33} \dot{q}_3 - g_3.
$$

Then, the following nonlinear state feedback

$$
\begin{aligned}
\tau_1 &= \frac{b_{21} b_{12} - b_{11} b_{22}}{b_{21}} u_2 + \frac{b_{21} b_{13} - b_{11} b_{23}}{b_{21}} u_3 + \\
&\quad + \frac{\gamma_1 b_{21} + \gamma_2 b_{11}}{b_{21}}, \\
f_3 &= \frac{b_{21} b_{32} - b_{31} b_{22}}{b_{21}} u_2 + \frac{b_{21} b_{33} - b_{31} b_{23}}{b_{21}} u_3 + \\
&\quad + \frac{\gamma_2 b_{31} - \gamma_3 b_{21}}{b_{21}},
\end{aligned}
$$

where $u_2, u_3 \in \mathbb{R}$ are two new inputs, transforms the system dynamics in:

$$
\begin{cases}
\ddot{q}_1 = -\frac{b_{23}}{b_{21}} u_3 + \frac{b_{22}}{b_{21}} u_2 - \frac{\gamma_2}{b_{21}}, \\
\ddot{q}_2 = u_2, \\
\ddot{q}_3 = u_3.
\end{cases}
$$

Furthermore, let us denote with $\hat{\zeta} = (\hat{q}_2, \hat{q}_3)^T$ a desired trajectory for output $\zeta$. Then, the following choice for new inputs $u_2$ and $u_3$ guarantees asymptotical convergence to zero of the tracking error $e_\zeta = \hat{\zeta} - \zeta = (\hat{q}_2 - q_2, \hat{q}_3 - q_3)^T$:

$$
\begin{aligned}
u_2 &= \ddot{\hat{q}}_2 + k_{v_2} \dot{e}_\zeta(1) + k_{p_2} e_\zeta(1), \\
u_3 &= \ddot{\hat{q}}_3 + k_{v_3} \dot{e}_\zeta(2) + k_{p_3} e_\zeta(2),
\end{aligned}
$$

where $k_{v_i}, k_{p_i} > 0$, for $i = 1, 2$.

Generating suitable desired trajectories for joint variables $q_2$ and $q_3$ is not trivial due to the system underactuation and the necessity of maintaining the tether cable in tension. Indeed, with the same spirit of [13], [14], two objectives have to be considered while designing $\hat{q}_2$ and $\hat{q}_3$: on the one hand, sufficient mechanical energy must be imparted to the end–effector during the startup phase so that it can reach the furthest point where a target–object can be placed, and on the other hand the tether cable must not become loose. Indeed, even the startup phase is composed of an initial step that aims at establishing a swing motion and a second step that allows maintenance of this motion [13], [14]. We will omit this for the sake of space. The reader may assume that, after the swing motion is established, its maintenance is achieved by imposing the following desired trajectories: $\hat{q}_2 = 0$, $\dot{\hat{q}}_2 = 0$, $\hat{q}_3 = \bar{q}_3$, and $\dot{\hat{q}}_3 = 0$. Therefore, in our case, the zero–dynamics reduces to the behavior of joint variable $q_1$, under the hypothesis that $q_2$ and $q_3$ have converged to their desired trajectories. At that time, we have $u_1 = u_2 = 0$, and the behavior of $(q_1, \dot{q}_1)$ reduces to the following differential equation:

$$
\ddot{q}_1 = -\frac{m_3 g \bar{q}_3}{I_3 + m_3 (\bar{q}_3^2 + a_1 \bar{q}_3)} \sin(q_1) = -\beta \sin(q_1), \quad (3)
$$

where $\beta$ is a constant depending on the acceleration gravity and the desired length of the tether cable, i.e. $\beta = \beta(g, \bar{q}_3)$. The asymptotic convergence of Eq. 3 was shown in [13], [14] but can be rigorously proved by applying Lyapunov direct method. Consider the following Lyapunov candidate function and restrict to consider the first joint $q_1$ in the set $Q_1 = (-\pi, \pi)$:

$$
V(q_1, \dot{q}_1) = \beta \left(1 - cos(q_1)\right) + \dot{q}_1^2, \quad (4)
$$

that is positive for all $q_1, \dot{q}_1 \neq 0$. Direct computation of its first time–derivative yields:

$$
\begin{aligned}
\dot{V}(q_1, \dot{q}_1) &= \beta \sin(q_1) \dot{q}_1 + 2 \dot{q}_1 \ddot{q}_1 = \\
&= \beta \sin(q_1) \dot{q}_1 - 2 \dot{q}_1 \beta \sin(q_1) = \quad (5) \\
&= -\beta \dot{q}_1 \sin(q_1) \leq 0.
\end{aligned}
$$

Moreover, the set $N = \{(q_1, \dot{q}_1)^T \mid \dot{V}(q_1, \dot{q}_1) = 0\}$ is given by all points where either $q_1 = \pi k$, for $k \in \mathbb{Z}$, or $\dot{q}_1 = 0$. Direct inspection of the system's trajectories that start from points in $N$ and that are complaint with the system dynamics of Eq. 3 and the condition $q_1 \in Q_1$ shows that the only stationary point for $V$ is $q_1 = 0, \dot{q}_1 = 0$, and hence the origin is asymptotically stable by Lasalle–Krasovskii principle.

Furthermore, let us denote with $p_e = (x_e, y_e, \phi_e)^T$ the configuration of the end–effector, where $(x_e, y_e)$ is the position of its center of mass, and $\varphi_e$ is its orientation w.r.t. a coordinate frame that is attached to the robot's base. The direct kinematics' relation, $p_e = k(q)$, can be easily found under the hypothesis that the robot is controlled so as to prevent the tether looseness. Indeed, we have:

$$
\begin{cases}
x_e = x_{base} + a_1 \tilde{S}_1 + q_3 \tilde{S}_{12}, \\
y_e = y_{base} - a_1 \tilde{C}_1 - q_3 \tilde{C}_{12}, \\
\varphi_e = q_1 + q_2 + \frac{\pi}{2},
\end{cases} \quad (6)
$$

where $(x_{base}, y_{base})$ is the position of the robot's base. Under the same hypothesis, the differential kinematics is easy to determine. Let us denote with $t_e = (\dot{x}_e, \dot{y}_e, \dot{\varphi}_e)^T$ the end–effector's twist vector, composed of its linear and angular velocities. This vector is related to the joint velocity $\dot{q} = (\dot{q}_1, \dot{q}_2, \dot{q}_3)^T$ by the following differential kinematics:

$$t_e = J(q)\,\dot{q}, \qquad (7)$$

where the manipulator Jacobian $J(q)$ is given by

$$J(q) = \begin{pmatrix} a_1\tilde{C}_1 + q_3\tilde{C}_{12} & q_3\tilde{C}_{12} & \tilde{S}_{12} \\ a_1\tilde{S}_1 + q_3\tilde{S}_{12} & q_3\tilde{S}_{12} & -\tilde{C}_{12} \\ 1 & 1 & 0 \end{pmatrix}.$$

It is straightforward to use the knowledge of direct and differential kinematics to computed the end–effector's state $\xi = (p_e^T, t_e^T)^T = (x_e, y_e, \phi_e, \dot{x}_e, \dot{y}_e, \dot{\phi}_e)^T$. Our choice is to have sensors measuring the joint position $q$ and velocity $\dot{q}$, whereas the vision system will be used only to detect the target–object position $p_t = (x_t, y_t)$.

### B. Real–time computation of end–effector's trajectory

In [6], the problem of steering a robotic end–effector toward a fixed object has been addressed. To catch objects placed at a generic position $p_t = (x_t, y_t)$ on the throwing plane, the end–effector's final state must be chosen as a function of $p_t$, i.e. $\hat{\xi}(t_f) = (x_t, y_t, \hat{\varphi}_e(p_t), \hat{\dot{x}}_e(p_t), \hat{\dot{y}}_e(p_t), \hat{\dot{\varphi}}_e(p_t))^T = \chi(p_t)$. Let us denote with $u$ a scalar control force transmitted through the tether cable. Then, we can readily pose the following real–time version of the problem:

*Problem 1 (Real–time steering problem):* Given an initial end–effector's state $\xi_0$ and a target position $p_t(t)$, find a feedback control strategy $\bar{u}(t) : [t_0, t_f] \to U$, where $t_0$ is the throwing time, $t_f$ is a suitable final time, and $U$ is the set of available controls, that steers the end–effector from $\xi_0$ to a suitable final state $\hat{\xi}(t_f) = \chi(p_t(t_f))$.

Previous work proposed a control strategy $u(t)$ to solve Problem 1, consisting of transmission of a series of impulses to the end–effector (see [9]). So–called multiple braking control steers an end–effector by means of a piecewise–constant function that is parameterized w.r.t. $r$ braking instants, $t_{b_i}$, and braking durations, $\Delta t_{b_i}$, i.e. $\bar{u}(t) = \lambda(t_{b_1}, \Delta t_{b_1}, \ldots, t_{b_r}, \Delta t_{b_r})^T$. Such a technique is effective and indeed very accurate, but applicable only for fixed objects, i.e. $p_t(t) = \bar{p}_t$, since computation of $\bar{u}(t)$ involves solving a nonlinear (nonconvex) system. A pictorial representation of the corresponding control scheme is reported in Fig. 3-a. More precisely, in the case of fixed objects, the constant position of the object, $\bar{p}_t$, can be initially measured, and then a suitable braking control $\bar{u}(t)$ can be computed *offline*. Application of this control to the controlled manipulator allows the end–effector's to reach the desired final state $\hat{\xi}(t_f)$.

Reaching or better yet catching moving objects requires closing a feedback loop of the current object position $p_t(t)$ on casting manipulation (see Fig. 3-b), but poses strict real–time specifications on the computation time of the inherent control strategy $\bar{u}(t) = \eta(\xi(t), p_t(t))$. This makes the *real–time casting manipulation problem*, including 3D control of the end–effector position, orientation, and velocity, a formidable challenge. Trying to solve the general problem, we restricted in [12] to controlling the end–effector's position by exploiting two simplified yet accurate models that are suitable for real–time computation, where the end–effector is replaced with a point of mass $m$.

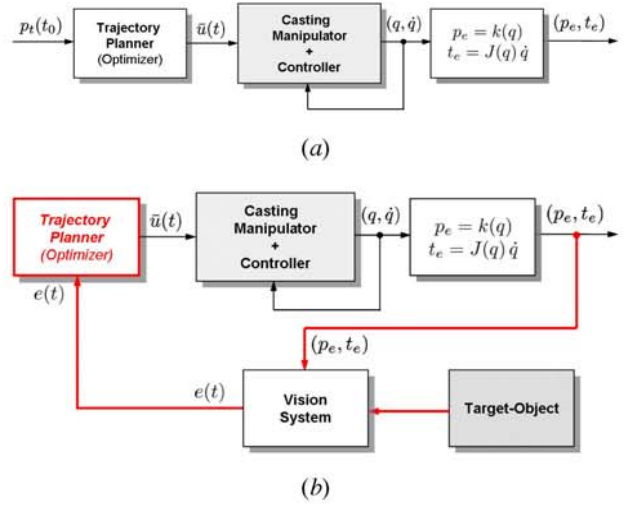

Figure 3. Offline (*a*) versus online (*b*) visual feedback of the target–object position

*1) Impulse–based control:* The first approach is simpler and consists of controlling the end–effector's landing position by exploiting one impulse. It is indeed a variation of the multiple braking control with only one braking time, and its steps are as follows. The target–object's position $p_t$ is first detected by the vision system, then the end–effector is imparted sufficient mechanical energy to reach a point further than $p_t$. The tether cable is released at a time $t_0$, and the end–effector is thrown. The $x$–coordinate of the furthest point that can be reached by the end–effector is obtained for $f_3(t) = 0$ for $t \geq t_0$ and is given by:

$$x_{land} = x_e(t_0) + \frac{\dot{x}_e(t_0)}{2g}\left(\dot{y}_e(t_0) + \sqrt{\dot{y}_e(t_0)^2 + 2g\,y_e(t_0)}\right),$$

where $g$ is the gravity acceleration, and the end–effector's initial state is obtained by evaluating Eq. 6 and 7. The *braking time* $t_b(t)$ is continuously updated by using the current object position $p_t(t)$, and it is computed as the instant when the length of the tether cable equals the distance between $p_t$ and the point where the tether departs from the robot ($H$ in Fig. 4). In formula we have:

$$\begin{aligned} ||(x_e(t_b), y_e(t_b))^T - (x_e(0), y_e(0))^T||_2 = \\ = ||(x_e(0), y_e(0))^T - (x_t(t), y_t(t))^T||_2\,, \end{aligned}$$

which gives an explicit feedback law for computing the braking time $t_b(t)$ based on the current target–object position $p_t$ and the end–effector's state $\xi(t)$:

$$t_b(t) = \chi(\xi(t), p_t(t))\,. \qquad (8)$$

At the braking time $t_b(t)$, either control input $f_3$ or a brake can be used to stop the tether unwinding and to fix its length, thus constraining the motion of the end–effector. A rough although effective approximation of the trajectory that the end–effector will perform is an arc of circumference $BC$ leading the end–effector directly over the desired target position $p_t$ (see Fig. 4). Since $t_b$ is computed based on the current measure of $p_t$, if the target–object comes nearer to the robot's base or recedes from it, the braking time is anticipated or delayed, respectively.

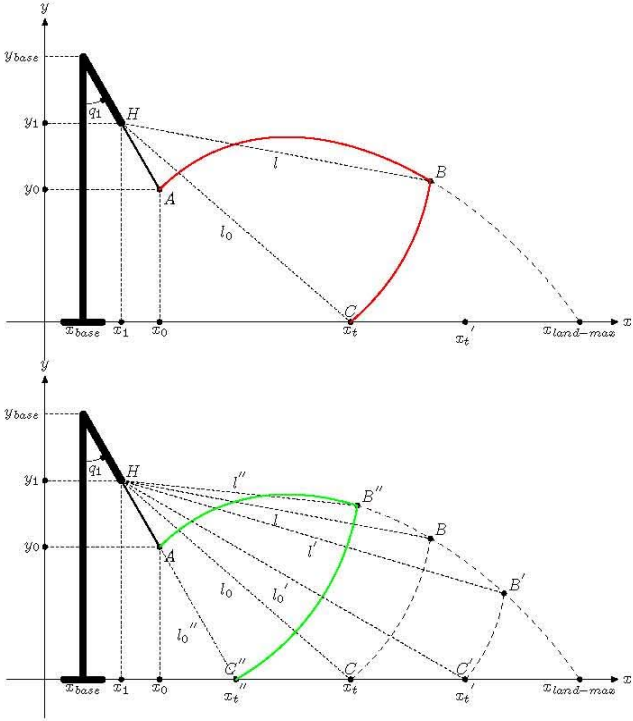

Figure 4. Depiction of possible end–effector's trajectories under impulse–based control method of the steering phase.

It is worth noting that this impulse–based approach is less sophisticated than the one of [9] for the fact that the elasticity of the tether cable is neglected and that the end–effector's orientation is discarded. However, the simplified problem allows us to compute $t_b$ as an explicit function of the target–object position. This makes it unnecessary the exploitation of iterative methods for its computation and then requires a negligible computation time for the control system. Furthermore, we use no information on the end–effector's velocity, although we assume it to be bounded. With the same spirit of [9] when using virtual points, we throw the end–effector to the furthest point that the target–object can reach. If the target–object moves further or continues to move after braking, the casting may fail, and the end–effector must be recast again.

*2) Piecewise–continuous force control:* A second more general approach involves transmission of a piecewise–continuous force signal $\bar{u}(t)$ to steer the end–effector during its flight. A simple model describing the end–effector's behavior under the hypothesis that the tether cable never becomes slack can be obtained from geometric considerations on Fig. 2:

$$\begin{cases} \ddot{x}_e = -\frac{1}{m}\sin\left[\tan^{-1}\left(\frac{x_e - x_{L_1}}{y_e - y_{L_1}}\right)\right] u, \\ \ddot{y}_e = -g - \frac{1}{m}\cos\left[\tan^{-1}\left(\frac{x_e - x_{L_1}}{y_e - y_{L_1}}\right)\right] u, \end{cases}$$

where $(x_{L_1}, y_{L_1})$ is the point where the tether departs from the rigid link. After some simplifications the end–effector's dynamics reduces to:

$$\begin{cases} \ddot{x}_e = -\frac{1}{m}\dfrac{\frac{x_e - x_{L_1}}{y_e - y_{L_1}}}{\sqrt{1+\left(\frac{x_e - x_{L_1}}{y_e - y_{L_1}}\right)^2}} u, \\ \ddot{y}_e = -g - \frac{1}{m}\dfrac{1}{\sqrt{1+\left(\frac{x_e - x_{L_1}}{y_e - y_{L_1}}\right)^2}} u. \end{cases} \quad (9)$$

Let us denote with $\xi = (x_e, y_e, \dot{x}_e, \dot{y}_e)^T$ the *reduced* end–effector's state (its orientation $\varphi_e$ and angular velocity $\dot{\varphi}_e$ are not relevant for our study). Let us also denote with $\xi_0$ the initial state at the throwing time $t_0$, and with $p_t$ the object current position. The end–effector's dynamics of Eq. 9 can be easily written in state form as $\dot{\xi}(t) = f(\xi(t), u(t))$. Denote also with $u_{max}$ the maximum force that can be transmitted through the cable. Then, the end–effector can be steered from state $\xi_0$ to the target–object in unknown time $t_f$, by a control function $\bar{u} : [t_0, t_f] \to [0, u_{max}]$ that solves the following dynamic programming problem:

$$\begin{cases} \bar{u}(t) = \arg\min_{u(t)\in U} J, \\ J = \|(x_e(t_f), y_e(t_f))^T - p_t\|_2^2, \\ \dot{\xi}(t) = f(\xi(t), u(t)), \\ \xi(t_0) = \xi_0, \\ t_f \geq 0, \\ 0 \leq u(t) \leq u_{max}. \end{cases} \quad (10)$$

To find an analytical solution of problem 10 is not affordable due to the fact that $f$ is nonlinear. Therefore, we look for approximations that can be obtained by using iterative numerical methods that are known in the research operation literature. These requires an initial guess of the solution function, $\bar{u}(t)^0$, and, at any step $k$, find better approximations $\bar{u}(t)^k$ by moving along the anti-gradient direction of a suitable objective function (see e.g. Newton, quasi–Newton methods). However, since $f$ is nonlinear, a bound on the number of steps that are needed to approximate the exact solution with a desirable accuracy is unknown. Indeed, knowledge of a "good" initial guess is essential to reduce the computation time and also to avoid local minima. Another factor that affects the time of convergence is the size of the *solution space* $U$, i.e. the class of functions where we look for a solution.

Therefore, we reduce the computation time and finally meet the real–time constraints by, on the one hand, limiting the solution space $U$, and, on the other hand, finding a number of good initial guesses $\bar{u}(t)^0$ that can be computed offline. More precisely, whenever a new measurement of the target–object's position $p_t(t_k)$ is available at a generic time $t_k$, we limit to seeking constant forces $\bar{u}(t) = \bar{u}(p_t(t_k)) = \bar{u}$ solving Problem 10. By doing this, we need to solve an optimization in only one variable. So–found constant function will be indeed applied up to next measurement at time $t_{k+1}$, thus obtaining an overall feedback control function that is a piecewise–constant, i.e. $\bar{u}(t) = u(\xi(t_0), p_t(t_0), \xi(t_1), p_t(t_1), \dots)$. A general expectation is that faster vision systems should allow us to compute piecewise–constant control function with smaller jumps at the switching times $t_k$. Furthermore, a *lookup table* of solution functions for a grid of points in the state space is prepared as a repository of good initial guesses $\bar{u}(t)^0$. Then, whenever a new measurement is available, only a small *variation* $\delta \bar{u}(t)$ from a nominal control signal $\bar{u}(t)^0$ has to be computed. By doing so, we reduce to solve a local variational formulation of Problem 10 that is *convex* w.r.t. $\delta \bar{u}$, and thus very efficient to solve. Indeed, computation time becomes negligible. Actual satisfaction of real–time constraints depends on the characteristic time of the end–effector's flight, which is very small, and the computation time of the feedback law. This aspect should be deeply considered, but goes beyond the scope of the work. Our purpose is to show that our technique is implementable on a reasonably common hardware (see Sec. IV). Fig. 5 is a depiction of a possible end–effector's trajectory under piecewise–constant control.

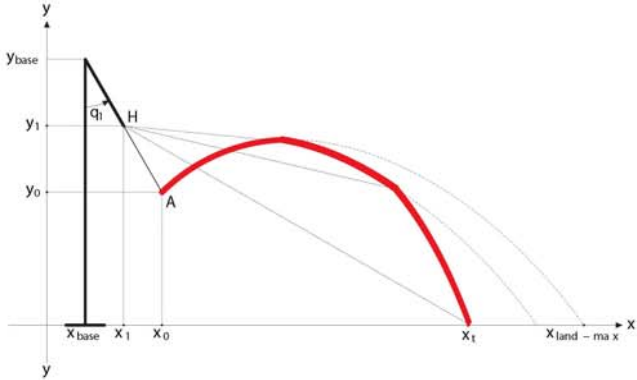

Figure 5. Depiction of possible end–effector's trajectories under piecewise–continuous control method of the steering phase.

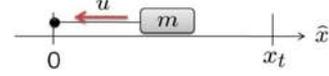

Figure 6. A prototypical example consisting of a tether cable attached to a mass $m$ and to the origin, that can exert a scalar force $u \in [0, u_{max}]$ on the mass itself.

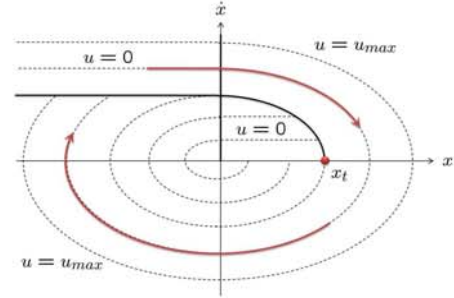

Figure 7. Optimal trajectories for a double integrator of mass $m$ with bounded input $u$. Input bounds are 0, and $u_{max}$.

In both approaches, joint torque $\tau_1$ is computed as in Sec. II so as to keep the first joint variable $q_1$ fixed to the throwing angle value. Experimental validation of the two approaches are reported in Sec. IV.

### III. 3-DIMENSIONAL CASTING MANIPULATION

This section expands on our work of [11], where the design of a novel mechanism for 3D casting was proposed. In this work, we first study a minimum–time control problem for a prototypical system with unilateral input control bounds, and then we deal with the 3D casting.

#### A. A prototypical 1–cable–driven mechanism

Consider the system of Fig. 6 consisting of a tether cable attached to a mass $m$ and to the origin, that can exert a scalar force $u \in [0, u_{max}]$ on the mass itself. The mass is constrained to move along a line. Let us denote with $x \in \mathbb{R}$ its coordinate. The system dynamics is described by the following equation:

$$m \ddot{x} = -\text{sign}(x)\, u\,. \tag{11}$$

Let us denote with $\xi = (x, \dot{x})$ the mass' state. We are then interested in solving the following:

*Problem 2:* (**minimum–time control of 1–cable–driven robot**) Find a control function $\bar{u}(t)$ that steers mass $m$ from an initial state $\xi_0 = (x_0, \dot{x}_0)$ toward a desired state $\xi_t = (x_t, 0)$, with $x_t \neq 0$, in minimum–time $t_f$.

Despite of the simplicity of the problem, we consider it for its substantial difference w.r.t. the classical example of a double integrator with symmetric input bounds, since in our case one bound is null. Formally, we have to solve the following dynamic optimization programme:

$$\begin{cases} \bar{u}(t) = \arg\min_{u(t)} J\,, \; J = \int_{t_0}^{t_f} 1\, d\tau = t_f - t_0\,, \\ \begin{pmatrix} \dot{\xi}_1(t) \\ \dot{\xi}_2(t) \end{pmatrix} = \begin{pmatrix} \xi_2(t) \\ -\frac{1}{m} \text{sign}(\xi_1(t))\, u(t) \end{pmatrix}, \\ \xi(t_0) = \xi_0\,, \; \xi(t_f) = \xi_t\,, \\ 0 \leq u(t) \leq u_{max}\,. \end{cases}$$

Solutions of our problem can be analytically computed by Pontryagin Minimum Principle [17] and are known to be bang–bang functions, i.e. $\bar{u}(t)$ is piecewise–constant and takes only value in $\{0, u_{max}\}$. Optimal trajectories of the mass are reported in Fig. 7 for different initial states. It can be shown that finite–time and finite–switching solutions exist for any case, except for $x_0 \neq 0$ and $x_t = 0$ that has no practical relevance.

#### B. Modeling and control of 3D casting manipulation

We are now ready to consider the 3D casting case. In [11], we proposed a robot consisting of a triangular platform rotating around a fixed axis, and 3 tether cables attached to the platform edges and to the end–effector. The rotation axis is orthogonal to the platform normal and passes through its center of mass. A depiction of the 3D casting robot is reported in Fig. 8.

The choice of having 3 tether cables in the mechanism is motivated by the fact that application of 3 independent forces maps, through the robot Jacobian, into a 3D force cone on the end–effector. Even when used for 2D casting, this allows exertion of a resultant force having non–null component orthogonal to the throwing plane. This ability can in fact be very useful whenever the end–effector leaves the throwing plane due to e.g. the presence of wind, or other disturbance.

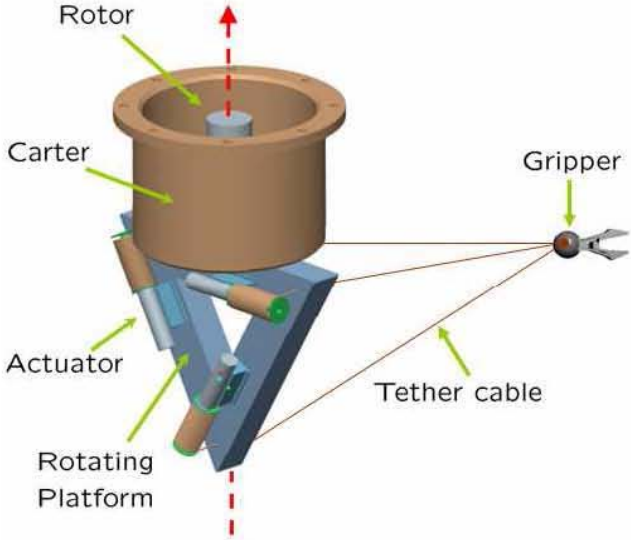

Figure 8. Depiction of a simple robot (DAVIDE) to be used for 3-dimensional casting manipulation.

Operation of the new manipulator is organized in the same phase as described in Sec. I. We focus here on the crucial steering phase for which we present a suitable model and derive the open–loop minimum–time control. We assume that a proper rotation of the platform is initially executed to impart the end–effector sufficient mechanical energy to reach the target–object. Then, the end–effector is thrown and the platform is fixed at the configuration that it possesses at the throwing time.

Modeling the system dynamics during the steering phase is easy if operating conditions prevent tether cable looseness, which we will assume. Let us again denote with $p_e = (x_e, y_e, z_e)$ the end–effector's position, with $u = (u_1, u_2, u_3)$ the actuation force, with $\alpha$ the angle between the platform and the target direction. Then the system dynamics reads:

$$m \begin{pmatrix} \ddot{x}_e \\ \ddot{y}_e \\ \ddot{z}_e - g \end{pmatrix} = -\Gamma(x_e, y_e, z_e, \alpha) \begin{pmatrix} u_1 \\ u_2 \\ u_3 \end{pmatrix}, \quad (12)$$
$$I \dot{\alpha} = I \omega = u_\alpha,$$

where

$$\Gamma = \begin{pmatrix} \frac{x_e + b \tilde{C}_\alpha}{\gamma_1(x_e,y_e,z_e)} & \frac{x_e - b \tilde{C}_\alpha}{\gamma_2(x_e,y_e,z_e)} & \frac{x_e}{\gamma_3(x_e,y_e,z_e)} \\ \frac{y_e + b \tilde{S}_\alpha}{\gamma_1(x_e,y_e,z_e)} & \frac{y_e - b \tilde{S}_\alpha}{\gamma_2(x_e,y_e,z_e)} & \frac{y_e}{\gamma_3(x_e,y_e,z_e)} \\ \frac{z_e + b \frac{\sqrt{3}}{3}}{\gamma_1(x_e,y_e,z_e)} & \frac{z_e + b \frac{\sqrt{3}}{3}}{\gamma_2(x_e,y_e,z_e)} & \frac{z_e + b \frac{\sqrt{3}}{3}}{\gamma_3(x_e,y_e,z_e)} \end{pmatrix},$$

$$\gamma_1 = \sqrt{\left(x_e + b \tilde{C}_\alpha\right)^2 + \left(y_e + b \tilde{S}_\alpha\right)^2 + \left(z_e + b \frac{\sqrt{3}}{3}\right)^2},$$
$$\gamma_2 = \sqrt{\left(x_e - b \tilde{C}_\alpha\right)^2 + \left(y_e - b \tilde{S}_\alpha\right)^2 + \left(z_e + b \frac{\sqrt{3}}{3}\right)^2},$$
$$\gamma_3 = \sqrt{x_e^2 + y_e^2 + \left(z_e + b \frac{2\sqrt{3}}{3}\right)^2},$$

$b$ is the length of the robot arm, and $I$ is the platform inertia. We have assumed that the dynamics of the rotating platform has been linearized, and that it appears to be linear w.r.t a new input $u_\alpha$.

We now focus on the mechanism control. The above study on the 1–cable–driven robot can be seen as a projection of our richer problem on one dimension, and it can give us some intuition on expected optimal trajectories for the robotic end–effector. Then, let us consider the problem of steering the end–effector toward a desired position in a 3D workspace. The problem is interesting due to the presence of unilateral bounds on inputs $u_1$, $u_2$, and $u_3$, reflecting the physics by which tether cables can exert only pulling forces. The optimization also concerns choice of the throwing angle Let us denote with $\xi = (x_e, y_e, z_e, \dot{x}_e, \dot{y}_e, \dot{z}_e)$ the end–effector's state, then its dynamics in Eq. 12 has the following state form $\dot{\xi}(t) = f(\xi(t), u(t), \alpha_0)$, where $\alpha_0$ is the throwing angle w.r.t. the target direction. Let us denote with $\omega_0$ the angular velocity of the rotating platform at the throwing time, then the initial state $\xi_0$ is obtained as:

$$\xi_0(\alpha_0, \omega_0) = (-r S_{\alpha_0}, r C_{\alpha_0}, 0, r \omega_0 C_{\alpha_0}, r \omega_0 S_{\alpha_0}, 0).$$

Denote with $x_f = (x_t, y_t, z_t)$ the target–object position, and with $u_{max}$ the maximum force that each actuator can provide. Then, the end–effector can be steered from state $\xi_0(\alpha_0)$ to $\xi_f = (x_f, 0)$ in unknown minimum time $t_f$, by finding the time–optimal control vector function $\bar{u} : [t_0, t_f] \to [0, u_{max}]$, and the optimal throwing angle $\bar{\alpha}_0 \in [0, 2\pi]$ that solve the following dynamic programming problem:

$$\begin{cases} [\bar{u}(t), \bar{\alpha}_0] = \arg\min_{[u(t),\alpha_0]} J, \; J = \int_{t_0}^{t_f} 1 \, d\tau = t_f - t_0, \\ \dot{\xi}(t) = f(\xi(t), u(t), \alpha_0), \\ \xi(t_0) = \xi_0(\alpha_0), \xi(t_f) = \xi_f, \\ 0 \leq u(t) \leq u_{max}. \end{cases}$$

We have solved this dynamic optimization problem by using numerical tools available in Matlab, and obtained the results that are summarized in the following figures. Fig. 9 shows the optimal time $\bar{t}_f$ versus the relative throwing angle $\alpha_0$, and reveals that the optimum value is $\bar{\alpha}_0 = 3\pi/4$ radiants. Fig. 10 shows the optimal time $\bar{t}_f$ at optimal throwing angle $\bar{\alpha}_0$ versus the target–object position: due to radial symmetry of the problem, optimal time depends only on the target distance. Furthermore, Fig. 11 and 12 show the optimal control $\bar{u}(t)$ for a point–to–point motion, and corresponding state evolution in condition of no gravity ($g = 0 \, m/s^2$), and with gravity ($g = 9.81 \, m/s^2$), respectively. In both simulations we assumed that actuators are such that is $u \in [-10, 10]$ N, and the mass of the end–effector is $1$ Kg. Throwing conditions in the simulations remain constant for the two case (with gravity and without gravity), although the final target–object positions changes since the shape of the workspace changes. This result is not reported and requires further study.

## IV. EXPERIMENTAL RESULTS OF VISUAL–BASED PLANAR CASTING MANIPULATION

Effectiveness of the two visual–based strategies proposed in Sec. II has been validated through experiments with a 2D

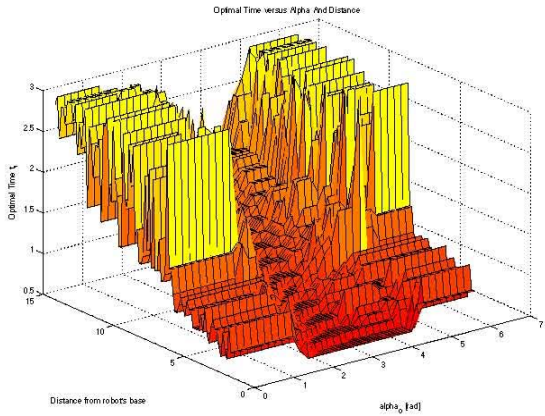

Figure 9. Optimal time $\bar{t}_f$ versus throwing angle $\alpha_0$ and target–object distance.

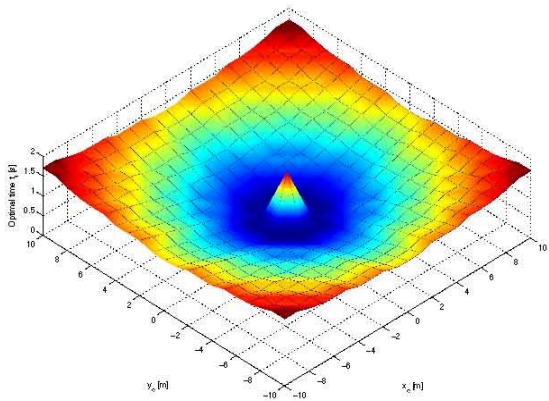

Figure 10. Optimal time $\bar{t}_f$ at optimal throwing angle $\bar{\alpha}_0$ versus the target–object position.

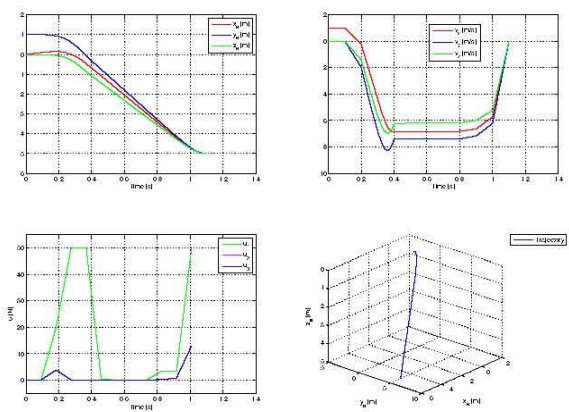

Figure 11. Optimal control $\bar{u}(t)$ for a point–to–point motion, and corresponding state evolution in condition of no gravity $(g = 0\,m/s^2)$.

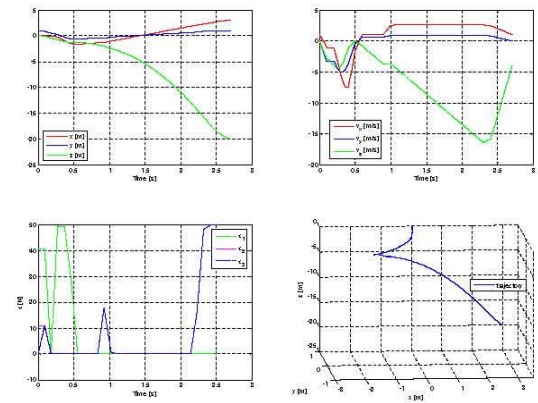

Figure 12. Optimal control $\bar{u}(t)$ for a point–to–point motion, and corresponding state evolution in the presence of gravity $(g = 9.81\,m/s^2)$.

| Parameter | Value | Unit |
|---|---|---|
| $x_{base}$ | 0.000 | $m$ |
| $y_{base}$ | 1.695 | $m$ |
| $a_1$ | 0.342 | $m$ |
| $m_1$ | 1.1105 | $kg$ |
| $I_1$ | 0.0216 | $kg\,m^2$ |
| $a_3$ | 0.495 | $m$ |
| $m_3$ | 0.084 | $kg$ |
| $I_3$ | 1.3440E-005 | $kg\,m^2$ |

Table II
INERTIAL AND GEOMETRICAL PARAMETERS OF THE REALIZED MANIPULATOR FOR PLANAR CASTING.

casting manipulator robot, under the same conditions as in simulation. We omit details on the identification of system parameters, but the reader may refer to [9] for an exhaustive description of this process.

### A. The robot

The manipulator realized in our lab is composed of a rigid link $L_1$ with revolute joint $q_1$ actuated by torque $\tau_1$, a revolute joint $q_2$, and a tether cable $L_3$ with translational joint $q_3$ actuated by input force $f_3$. Since we restricted to consider position control of the end–effector, the gripper has been replaced with a point of mass $m_3$. Refer to Sec. II for the robot's and the end–effector's dynamics. Table II reports the robot's geometrical and inertial parameters.

Optical encoders are used to measure joint variables q. An 81 000–pulses–per–revolution (ppr) encoder is used for the first joint, and two 2048–ppr encoders for the other joints. Two direct–drive motors are used to apply input controls $\tau_1$ and $f_3$. To control the tether unwinding control input $f_3$ is used as well as a braking mechanism. Fig. 13 shown the robot and a detail of the braking mechanism.

### B. Vision and control systems

A vision system can be used to measure position of the target–object as well as position and orientation of the end–effector during the ballistic flight phase. To this purpose, high–speed cameras achieving sampling times around 1 msec can be exploited. However, a very high framerate such as the one used in [18], [19] imposes a very high computation load. By contrast, to meet real–time constraints, we restrict to use the vision system only for detecting the position $(x_t, y_t)$ of the target–object. Furthermore, we assumed that the target–object at a maximum speed of 0.5 m/sec for which a framerate of 62 msec for the vision system leads to sufficiently accurate measures. We examined the accuracy of the vision system by conducting basic experiments of measuring the position of a target–object oscillating around a given position with maximum speed limited to 0.5 m/sec.

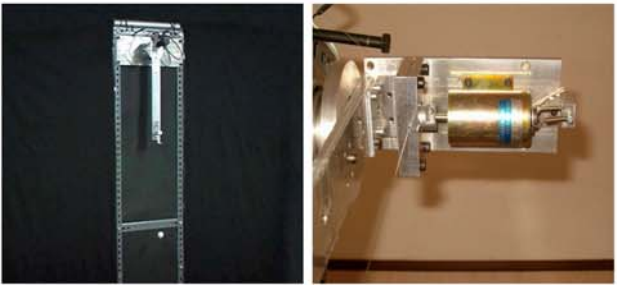

Figure 13. The manipulator realized in our lab, and a detail of the brake.

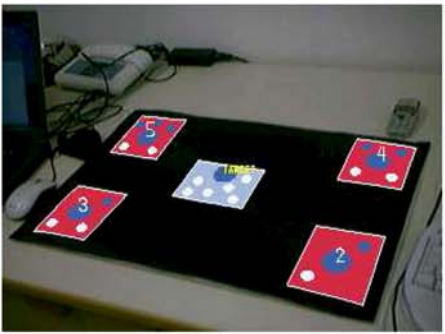

Figure 14. Example of image processed by the vision system with detected items: the target–object is highlighted in blue, and other calibration markers are highlighted in red.

Position $p^{img}$ of an object detected by the vision system on the captured frame has to be converted into real–world coordinates $p^{real}$. This is achieved by definition of a so–called *homography map* $\mathcal{H}$ such that $p^{real} = \mathcal{H}(p^{img})$. Choice of a suitable map $\mathcal{H}$ depends on hypotheses on the considered environment, and tuning of its inherent parameters can be cumbersome. In our scenario, we restrict to consider objects moving on a fixed plane for which a linear transformation $p^{real} = H p^{img}$ is sufficiently accurate, as it is shown in [20]. Estimation of parameters in H representing the camera calibration is straightforward, and can be found e.g. in [20], thus it is omitted here for the sake of space. Furthermore, extraction of the target–object's position $p_t^{img}$ on the captured frame is based a *contour–detection* algorithm. Fig. 14 shows an example of image processed by the vision system with all detected objects.

The vision system is easy to set up, and hard to fail in tracking the target. After calibration, the vision system achieves an accuracy of $4 \cdot 10^{-3}$m with a camera placed at around 3m in front of the throwing plane, thus showing that the linearity assumption for the homography map is valid for the considered scenario. The computation load for the extraction of the target–object position, and even the calibration are very low and meet the real–time constraints. This allows re–calibrations during the steering phase.

The robot control and the vision systems always runs concurrently with higher priorities than all other processes activates in the operating system. They run on a single Pentium© IV with clock frequency of 3.0 GHz. A US Digital PCI4ES card are used to read the encoders measuring the joints' positions q and velocities $\dot{q}$, and a National Instrument PCI6024E card is used to send the inputs $(\tau_1, f_3)$ to the robot. The vision system uses a USB Logitech© Orbit camera with frame–rate of 16 fps. The software architecture runs on a Microsoft Windows© XP platform. The source code for control and vision was written in C++ programming language. Moreover, implementation of the algorithms for calibrating the camera and extracting the target–object position exploit the OpenCV© library. Scheduling period for the control process is 0.5 msec.

### C. Experiments

First we report results validating the impulse–based control strategy. We run the following type of experiment. As soon as the system is turned on, the target–object position $p_t$ is measured, and the end–effector is imparted sufficient mechanical energy as described in Sec. II (startup phase). When a suitable throwing condition is achieved, the end–effector is cast, and the tether starts unwinding. In the first part of the experiment, we keep the position of the object constant, $p_t(t) = \bar{p}_t$. At time $t_b(t_0)$, when the tether length $q_3$ equals the distance between the target position and the point where the tether departs from the first link, the unwinding is stopped. As expected from theory, the end–effector's ballistic flight is composed of an arc of parabola and an arc of circumference corresponding to curves $AB$ and $BC$ of the figure, respectively, and the object is reached. As a second part of the experiment, the target–object is placed at the same position. As a soon as the robot throws its end–effector, the target–object is moved to a new position that is closer to the robot's base (let us say at time $t_b(t_1)$). As expected from theory, the target is reached by braking the tether unwinding at a smaller braking time, i.e. $t_b(t_1) < t_b(t_0)$. An example of the end–effector's trajectory under impulse–based control is reported in Fig. 15, and relevant signals are shown in Fig. 16.

Secondly, we report results validating the piecewise–constant force controller. The same type of experiment has been run with a target–object moving closer to the robot's base at a maximum speed of 0.5 m/sec. The robot is turned on at time $t_{-1} = 0$ sec and the target–object position is detected at 1.81 m from the robot's base. The end–effector is imparted sufficient mechanical energy (startup phase) and is thrown at time $t_0 = 8.461$ sec. The end–effector is steered while new measurements of the target–object position are available at time $t_1 = 8.710$ sec, $t_2 = 8.967$ sec, and $t_3 = 9.255$ sec. The overall steering phase concludes at time $t_f = 9.6105$ sec.

In our implementation, the optimization procedure has been written in C/C++ for achieving better efficiency and for full integration with the control and vision systems. Control variations $\bar{\delta}u(t)$ need calculation times much smaller than the vision system rate, and it is compliant with the target–object maximum velocity. Hence it can be performed online. An example of trajectory executed by the system under this control strategy is reported in Fig. 17, and the corresponding relevant signals are shown in Fig. 18.

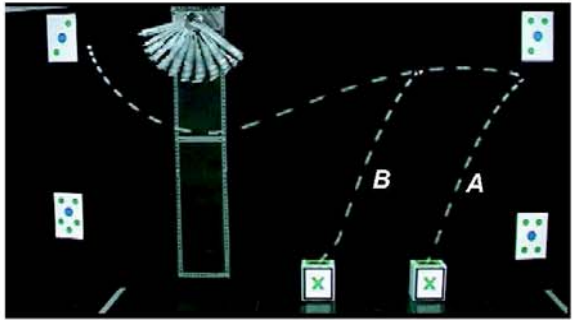

Figure 15. An example of the end–effector's trajectory under the impulse–based control method for the planar casting manipulator. Trajectories $A$ and $B$ show how an object coming closer to the robot's base can be reached.

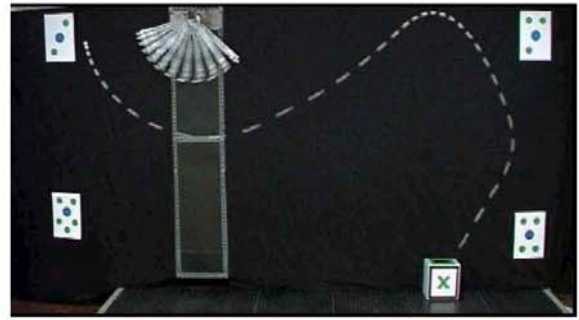

Figure 17. An example of the end–effector's trajectory under the continuous control method for the planar casting manipulator.

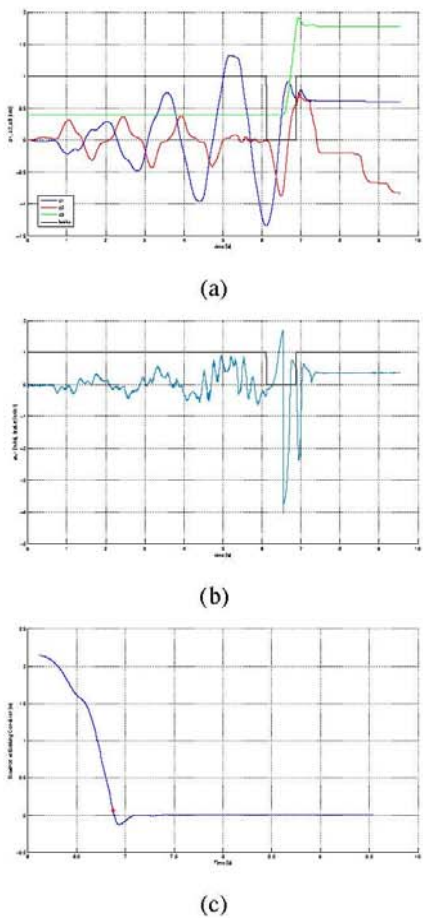

Figure 16. Data from an experiment with the impulse–based control scheme: joint variables (a), control and brake activation (b), and distance from the braking condition (c).

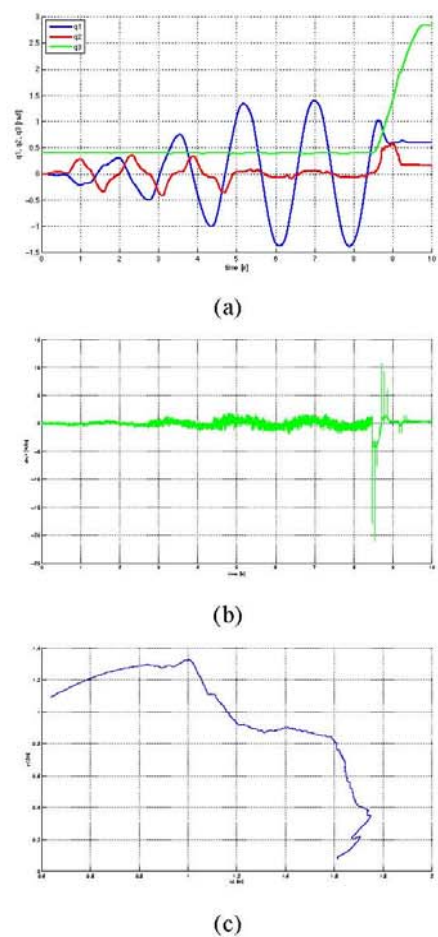

Figure 18. Data from an experiment with the continuous control method: joint variables (a), control (b), and end–effector's trajectory after the start–up phase (c).

Despite the very low computation time required by the impulse–based control, this method does not allow to reach target–objects moving after the braking time, except for particular cases. Moreover, as it can be seen in Fig. 15, actual trajectories of the end–effector after the braking can only be very roughly approximated by arcs of circumference, thus showing how the modeling hypothesis of tether tautness becomes weak after that moment. Although a volume of data sufficient to statistically compare the success rate of the two methods has not been collected, it appeared empirically that the continuous control method was more reliable. Indeed, the obtained trajectories were smoother — thus reducing the tether wear —, and the tether remained in tension during the flight phase, thus keeping the modeling hypothesis still valid. By contrast, the computational load of the continuous control method is not negligible, and strict requirements on computing hardware have to be considered to meet the real–time constraints.

## V. Conclusion

In this article we addressed the problem of catching objects that move at a relatively large distance from the robot. To this purpose, we adopt casting manipulation and visual–based feedback control. In previous work, planar casting manipulation was demonstrated to aptly catch static objects placed at a distant, known position, thus proving it viable for applications such as sample acquisition and return, rescue, etc. In this paper we proposed an extension of the idea to controlling the position of the end-effector to reach unpredictably moving targets in 3D. The goal has been achieved by an innovative design of the casting mechanism, and by closing a real–time control loop on casting manipulation using visual feedback of moving targets. Simplified yet accurate models of the system suitable for real–time computation have been developed, along with a suitable visual feedback scheme for the flight phase. Effectiveness of the visual feedback controller was demonstrated through experiments with a 2D casting robot. Future work will concern 3D control of the end–effector orientation.

## Acknowledgment

The work has been done with partial support by EC through project PHRIENDS (Contract IST-2006-045359).

## References

[1] O. Khatib, K. Yokoi, K. Chang, D. Ruspini, R. Holmberg, and A. Casal, "Vehicle/arm coordination and multiple mobile manipulator decentralized cooperation," *Proc. IEEE International Conference on Intelligent Robots and Systems*, 1996.

[2] R. Mamen, "Applying space technologies for human benefit: the canadian experience and global trends," *Canadian Space Agency*, 1986.

[3] H. Mochiyama, E. Shimemura, and H. Kobayashi, "Shape correspondence between a spatial curve and a manipulator with hyper degrees of freedom," *Proc. IEEE International Conference on Intelligent Robots and Systems*, 1998.

[4] N. Takanashi, H. Chose, and J. Burdick, "Simulated and experimental results of dual resolution sensor based planning for hyper redundant manipulators," *Proc. IEEE International Conference on Intelligent Robots and Systems*, 1993.

[5] H. Arisumi, T. Kotoku, and K. Komoriya, "Swing motion control of casting manipulation," *IEEE Control Systems*, 1999.

[6] H. Arisumi and K. Komoriya, "Posture control of casting manipulation," *Proc. IEEE International Conference on Robotics and Automation*, 1999.

[7] ——, "Study on casting manipulation (midair control of gripper by impulsive force)," *Proc. IEEE International Conference on Intelligent Robots and Systems*, 1999.

[8] ——, "Catching motion of casting manipulator," *Proc. IEEE International Conference on Intelligent Robots and Systems*, 2000.

[9] H. Arisumi, K. Yokoi, and K. Komoriya, "Casting manipulation - midair control of gripper by impulsive force," *IEEE Trans. on Robotics*, May, 2008.

[10] M. Rahimi, H. Hemami, and Y. Zheng, "Experimental study of a cable-driven suspended platform," *Proc. IEEE International Conference on Robotics and Automation*, vol. 3, 1999.

[11] A. Fagiolini, A. Torelli, and A. Bicchi, "Casting robotic end–effectors to reach far objects in space and planetary missions," *Proc. of 9th ESA Workshop on Advanced Space Technologies for Robotics and Automation*, 2006.

[12] A. Fagiolini, H. Arisumi, and A. Bicchi, "Visual-based feedback control of casting manipulation," *Proc. IEEE International Conference on Robotics and Automation*, pp. 2203–2208, 2005.

[13] H. Arisumi, T. Kotoku, and K. Komoriya, "A study of casting manipulation (swing motion control and planning of throwing motion)," *Proc. IEEE International Conference on Robotics and Automation*, 1997.

[14] ——, "Swing motion control of casting manipulation (experiment of swing motion control)," *Proc. IEEE International Conference on Robotics and Automation*, 1998.

[15] ——, "Study on casting manipulation (experiment of swing control and throwing)," *Proc. IEEE International Conference on Intelligent Robots and Systems*, 1998.

[16] A. D. Luca and G. Oriolo, "Motion planning under gravity for underactuated three-link robots," *Intelligent Robots and Systems*, 2000.

[17] L. Pontryagin, V. Boltyanskii, R. Gamkrelidze, and E. Mishchenko, "The Mathematical Theory of Optimal Processes," *Interscience, New York*, 1962.

[18] U. Frese, B. Bauml, S. Haidacher, G. Schreiber, I. Schaefer, M. Hahnle, and G. Hirzinger, "Off-the-shelf vision for a robotic ball catcher," *Proc. IEEE International Conference on Intelligent Robots and Systems*, vol. 3, pp. 1623–1629, 2001.

[19] A. Namiki, T. Komuro, and M. Ishikawa, "High-speed sensory-motor fusion for robotic grasping," *Measurement Science and Technology*, vol. 13, 2002.

[20] R. Hartley and A. Zisserman, "Multiple view geometry in computer vision," *Proc. IEEE International Conference on Intelligent Robots and Systems*, 1999.